\documentclass{article}

\usepackage{arxiv}
\usepackage{natbib}
\usepackage[utf8]{inputenc} % allow utf-8 input
\usepackage[T1]{fontenc}    % use 8-bit T1 fonts
\usepackage{hyperref}       % hyperlinks
\usepackage{url}            % simple URL typesetting
\usepackage{booktabs}       % professional-quality tables
\usepackage{amsfonts}       % blackboard math symbols
\usepackage{nicefrac}       % compact symbols for 1/2, etc.
\usepackage{microtype}      % microtypography
\usepackage{lipsum}
\usepackage{graphicx}
\graphicspath{ {./images/} }
\usepackage[nameinlink,noabbrev,capitalise,sort]{cleveref}
\usepackage{todonotes}

\usepackage{pifont}%
\newcommand{\cmark}{\ding{51}}%
\newcommand{\xmark}{\ding{55}}%

\usepackage{enumitem}

\title{Towards Understanding Visual Grounding in Vision-Language Models}

\author{
 Georgios Pantazopoulos \\
  The Alan Turing Institute\\
  Heriot-Watt University\\
   \And
  Eda B. \"Ozyi\u{g}it \\%\thanks{Contact: eozyigit@turing.ac.uk} \\
  The Alan Turing Institute\\
}
\renewcommand{\today}{}
\begin{document}
\maketitle
\begin{abstract}
Visual grounding refers to the ability of a model to identify a region within some visual input that matches a textual description.
Consequently, a model equipped with visual grounding capabilities can target a wide range of applications in various domains, including referring expression comprehension, answering questions pertinent to fine-grained details in images or videos, captioning visual context by explicitly referring to entities, as well as low- and high-level control in simulated and real environments.
In this survey paper, we review representative works across the key areas of research on modern general-purpose vision-language models (VLMs). We first outline the importance of grounding in VLMs, then delineate the core components of the contemporary paradigm for developing grounded models, and examine their practical applications, including benchmarks and evaluation metrics for grounded multimodal generation. We also discuss the multifaceted interrelations among visual grounding, multimodal chain-of-thought, and reasoning in VLMs. Finally, we analyse the challenges inherent to visual grounding and suggest promising directions for future research.
\end{abstract}

% keywords can be removed

\section{Introduction}
Visual grounding \cite{li2etal025} represents a fundamental challenge in multimodal artificial intelligence, requiring systems to establish precise correspondences between natural language descriptions and specific regions or objects within visual scenes. 
This capability is essential for enabling machines to comprehend and manipulate visual content through linguistic interaction, forming the cornerstone of human-computer interfaces that rely on natural language communication about visual information. 
The field encompasses a diverse spectrum of tasks and applications, ranging from core problems such as referring expression comprehension \citep{kazemzadeh-etal-2014-referitgame, he2023grec, krishna2017visual} and grounded captioning \citep{PontTuset_eccv2020, peng2023kosmos} to complex applications including grounded visual question answering \citep{zhu2016visual7w, lai2024lisa}, low- and high-level robotic control in interactive environments \citep{shridhar2020alfred, jiang2022vima, gao2023alexa}, and agents operating across web, mobile, and desktop environments \citep{cheng2024seeclick, gou2025navigating}.

Visual grounding has evolved through several distinct paradigms. Early approaches combined Convolutional Neural Networks (CNNs) with Recurrent Neural Networks (RNNs) \citep{Mao_2016_CVPR, Liu_2017_ICCV, yu2016modeling}. These were followed by specialised transformer-based models emerging from multi-task vision language pre-training (VLP) \citep{kamath2021mdetr, deng2021transvg, chen2020uniter, wang2022ofa, cho2021unifying, pantazopoulos-etal-2023-multitask}, and more recently by methods that leverage vision-language models (VLMs) \citep{youferret, chen2023shikra, bai2023qwen, bai2025qwen2}.
This progression toward grounded text generation has enabled multimodal models to establish intrinsic connections between linguistic expressions and corresponding visual elements, equipping machines with increasingly sophisticated fine-grained multimodal comprehension capabilities.

This survey primarily examines multimodal grounded generation in VLMs published from 2022 onward, while also incorporating key insights from foundational VLP models. The main contributions of this work are: (i) a comprehensive overview of grounding in VLMs, extending beyond the traditional focus on generalised Referring Expression Comprehension (REC) \citep{kazemzadeh2014referitgame} to encompass underexplored research domains; (ii) an analysis of how core architectural components influence grounding capabilities, offering a perspective absent from prior surveys \citep{liang2024survey, wu2023multimodal, huang2023visual, caffagni2024revolution}; (iii) a coverage of additional grounding-related tasks, such as grounded captioning, grounded visual question answering, and agents interacting with Graphical User Interfaces (GUIs); and (iv) a discussion of grounding as a component of multimodal reasoning, both as an intermediate step and as part of solution-level reasoning processes.

%\begin{itemize}[leftmargin=*]
%\item  A comprehensive overview of grounding in VLMs, extending beyond the traditional focus on generalised Referring Expression Comprehension (REC) \citep{kazemzadeh2014referitgame} to encompass underexplored research domains.

%\item An analysis of how core architectural components influence grounding capabilities, offering a perspective absent from prior surveys \citep{liang2024survey, wu2023multimodal, huang2023visual, caffagni2024revolution}.

%\item Coverage of additional grounding-related tasks, such as grounded captioning, grounded visual question answering, and agents interacting with Graphical User Interfaces (GUIs).

%\item A discussion of grounding as a component of multimodal reasoning, both as an intermediate step and as part of solution-level reasoning processes.

%\end{itemize}

The remainder of this paper is organised as follows. \cref{sec:background} presents the essential background, including definitions and relevant research domains. \cref{sec:grounding_vlms} synthesises prior work on the development of VLMs with grounding capabilities. \cref{sec:approaches} provides a comprehensive analysis of existing applications, datasets, and benchmarks, spanning image-level visual grounding to agents that interact with environments and web interfaces. Finally, \cref{sec:challenges} discusses the key components involved in developing grounded VLMs and examines the challenges and open questions facing the next generation of these models.

\section{Preliminaries}\label{sec:background}
%\todo[inline]{Add a main figure with different tasks?}

\subsection{What is Visual Grounding?}
In earlier works, visual grounding, where a model is tasked with localising a specific region within a visual context that aligns with an input textual query \citep{xiao2024towards}, is often referred to as REC \citep{kazemzadeh2014referitgame} or phrase localisation (PL) \citep{plummer2015flickr30k}, where the input query describes exactly one entity in an image. However, this definition is narrow and does not adequately encompass neighbouring fields such as grounded captioning, visual question answering grounded in visual elements, and interactive agent-based environments. In these settings, the textual query may take various forms, for instance, specifying a region of interest (e.g. describe the region with a concise caption), posing a question (e.g. which of these objects is used to cut items?), or giving an instruction (e.g. clean up old cookies from the Amazon store).  Accordingly, we define visual grounding more broadly as the capability to precisely localise and identify specific objects, regions, or concepts within a visual context, conditioned on natural language descriptions or instructions, thereby bridging textual understanding and visual reasoning. 

\subsection{The Role of Grounding in Modern VLMs}

Grounding in VLMs is essential not only for enabling fine-grained downstream applications, but also for establishing an interpretable link between predicted perceptions and actual localisation within the visual context. Although VLM outputs often imply an understanding of visual elements, grounding explicitly verifies whether the model can accurately identify and localise the relevant objects, regions, or concepts. This connection between language and precise visual evidence is critical for both transparency and reliability in multimodal reasoning. Prior research indicates that hallucination in VLMs arises from an excessive reliance on the language prior, with the dependence on the visual prompt diminishing as more tokens are generated \citep{favero2024multi, goyal2017making, wang2023evaluation, frank-etal-2021-vision, lin2023revisiting}. Without proper grounding, these models may generate plausible-sounding descriptions that do not correspond to actual visual content, resulting in unreliable outputs in critical applications. Grounding addresses this issue by ensuring that when a model describes an object or relationship, it can demonstrate its understanding by precisely identifying where that information appears in the visual input. 

Consequently, VLMs without grounding capabilities often struggle to establish precise correspondences between textual descriptions and visual elements. This can result in responses that are semantically plausible but spatially inaccurate. Recent research \citep{youferret, chen2023shikra} suggests that grounding mitigates this issue by encouraging models to commit to specific spatial locations, thereby enhancing both the accuracy and interpretability of their visual understanding.
However, subsequent studies \citep{lin2023revisiting} have shown that grounding objectives do not exhibit a causal relationship with reductions in object hallucinations, as quantified in visual question answering and image captioning tasks, indicating that further investigation is necessary. Nevertheless, the ability to precisely refer to regions that align with open-ended textual input remains an invaluable asset, enabling VLMs to support a wide range of downstream applications \citep{kazemzadeh-etal-2014-referitgame, zhu2016visual7w, zheng2024seeact, wu2023webui}.

\subsection{Representing Visual References}
Visual grounding requires converting continuous spatial coordinates into formats that language models can process effectively. 
Consequently, the design choice of representing regions plays a crucial role in model development. There are two broad categories for representing the bounding box: object-centric and pixel-level.

\paragraph{Object-centric} Object-centric methods present the model with region proposals, optionally accompanied by image-level features, and task the model with selecting the candidate proposal that satisfies the input textual query. In this framework, each individual object is processed separately by a vision component \citep{Girshick_2015_ICCV, ren2015faster, He_2017_ICCV, reis2023real, khanam2024yolov11}. Consequently, these methods offload visual perception to a unimodal expert and produce models that learn to match textual queries to one or more object candidates provided as inputs. The matching is typically implemented using sentinel tokens \citep{cho2021unifying, pantazopoulos-etal-2023-multitask, zhang2025gpt4roi}, where each sentinel token corresponds to the position of a bounding box in the sequence. Such a design offers interpretable object-level reasoning and a clean representation for detection and grounding tasks. However, it is constrained by predefined object categories, a lack of fine-grained spatial relationships, and the computational overhead from multi-stage processing. These limitations, combined with the flexibility and capacity of ViT-based encoders, have led to the prevalence of models that adopt a pixel-level approach.

\paragraph{Pixel-level} In pixel-level approaches, the image is divided into a grid of patches, typically using a vision transformer (ViT) \citep{zhai2023sigmoid, radford2021learning, tschannen2025siglip, carion2020end, zhang2021vinvl} pretrained with contrastive objectives on image-text pairs. 
The ViT \citep{dosovitskiyimage} partitions the image into a fixed-size grid of patches (e.g. a patch may represent $14\times14$, $16\times16$, $32\times32$ pixels), with each patch embedded as a visual token that encodes localised information. 
Pixel-level visual regions are commonly represented in two ways: discretising the image into a grid with fixed-size bins or directly encoding coordinates as textual outputs. The choice between discretised and raw coordinate representations has significant implications for model performance and capability. Discretised coordinates involve rounding continuous coordinates to discrete levels \citep{wang2022ofa, luunified2023, yang2022unitab, wang2023visionllm, wanggit}, thereby reducing the vocabulary size while maintaining reasonable spatial precision. 
The quantisation level, thus, serves as a critical hyperparameter, balancing accuracy, and computational efficiency. As an alternative, vector quantisation obtains patch representations from the latent space of an autoencoder, yielding a learned set of discrete visual tokens \citep{bigverdi2025perception}. In contrast, raw numeric tokens encode region coordinates directly as digit sequences, analogous to specifying a bounding box numerically such as \( \langle 142, 67, 298, 134 \rangle \) \citep{youferret, chen2023shikra, bai2025qwen2, wang2024cogvlm, chen2024expanding}. While this method offers higher spatial precision, it requires the model to perform arithmetic and compositional reasoning over tokenised numbers, which poses a challenge for transformer architectures not inherently designed for numerical reasoning.

\paragraph{Object-centric vs Pixel-level} The representation of regions of interest is a key component of model design. While earlier methods were largely object-centric, contemporary vision-language models predominantly adopt pixel-level representations with vision transformer backbones. Accordingly, the remainder of this survey focuses on pixel-level approaches.

\paragraph{Discretised vs Raw coordinates} Recent work has investigated the choice between discretised and raw coordinate formats at the pixel-level, indicating that the raw-coordinate format may be advantageous over discretisation \citep{chen2023shikra}. However, an in-depth comparison between the two has yet to be conducted. A possible explanation for the observed advantage of the raw-coordinate format is that, during the development of the VLM, the model already acquires some knowledge of numerical order, likely as a result of pre-training the language backbone on text data. In contrast, the embeddings of discrete tokens are typically initialised from scratch, requiring the model to learn their meanings and mappings without any prior knowledge. Nevertheless, the raw-coordinate format results in shared representations for numerical characters, which may lead to ambiguity. For example, a general-purpose VLM capable of both visual grounding and visual question answering may assign the same learned representation to the token ``$3$'', regardless of its visual or textual context. This phenomenon, often referred to as the \textit{modality gap} \citep{NEURIPS2022_702f4db7}, can significantly impact downstream performance.

\paragraph{Set-of-Marks} An alternative way of representing visual references in VLMs is through the use of a Set-of-Marks (SOM) \citep{yang2023set, yang2023dawn, wan2024contrastive, yang2023fine, lei2025scaffolding, cai2024vip}. 
This method follows a two-step process and it is particularly suitable to VLMs that are not equipped with grounding capabilities during training. In the first step, the image is partitioned into regions with off-the-shelf segmentation models \citep{ravi2024sam, kirillov2023segment, zou2023segment}. These regions are then labeled with a set of marks such as alphanumeric characters, symbols, masks, boxes, or colors. In the second step, the VLM performs visual grounding by leveraging the marks depicted in the image. It is worth noting that, while this approach can be effective, the performance of the VLM is strongly dependent on its Optical Character Recognition (OCR) and reading comprehension capabilities.

These representations underpin the grounding process, enabling models to align textual queries with corresponding visual regions. Framing grounding as a cross-modal retrieval objective provides a coherent perspective for understanding how these components contribute to effective multimodal reasoning.

\subsection{Grounding as an In-Context Cross-Modal Retrieval Task}

Visual grounding can be framed as an in-context multimodal retrieval task in modern VLMs \citep{pantazopoulos-etal-2024-shaking}.
Both tasks follow a similar pattern: given a context and a query, the model is required to locate and reference the relevant part of the context. 

In standard text-based in-context retrieval \citep{rajpurkar-etal-2018-know}, the model receives a passage of text (the context) and a question (the query), and it is tasked with identifying and extracting the specific text span that answers the question. This grounding task follows the same structure but operates across modalities. The model receives visual patches as the context and a textual description as the query, and it aims to  determine which visual region corresponds to the description.

The key distinction lies in the embedding spaces and the requirement for cross-modal mapping. In text-based retrieval, both input and output reside in the same semantic space (text tokens to text tokens). Visual grounding, by contrast, requires the model to bridge two distinct representation spaces. The model is required to understand the semantic content of the textual query, map that understanding onto the visual patch representations, and then translate the result back into textual coordinates or descriptions. Viewed from this perspective, the VLM performs a two-step process: first matching the textual query to the visual modality and then generating a textual response. Aligning text and visual representations remains the core difficulty of visual grounding, underscoring the need for effective cross-modal reasoning.

%This framing important as it illustrates why some architectural choices are better than others for VLMs (see \cref{sec:grounding_mllms}).

\begin{figure*}[ht!]
    \centering
    \includegraphics[width=\linewidth]{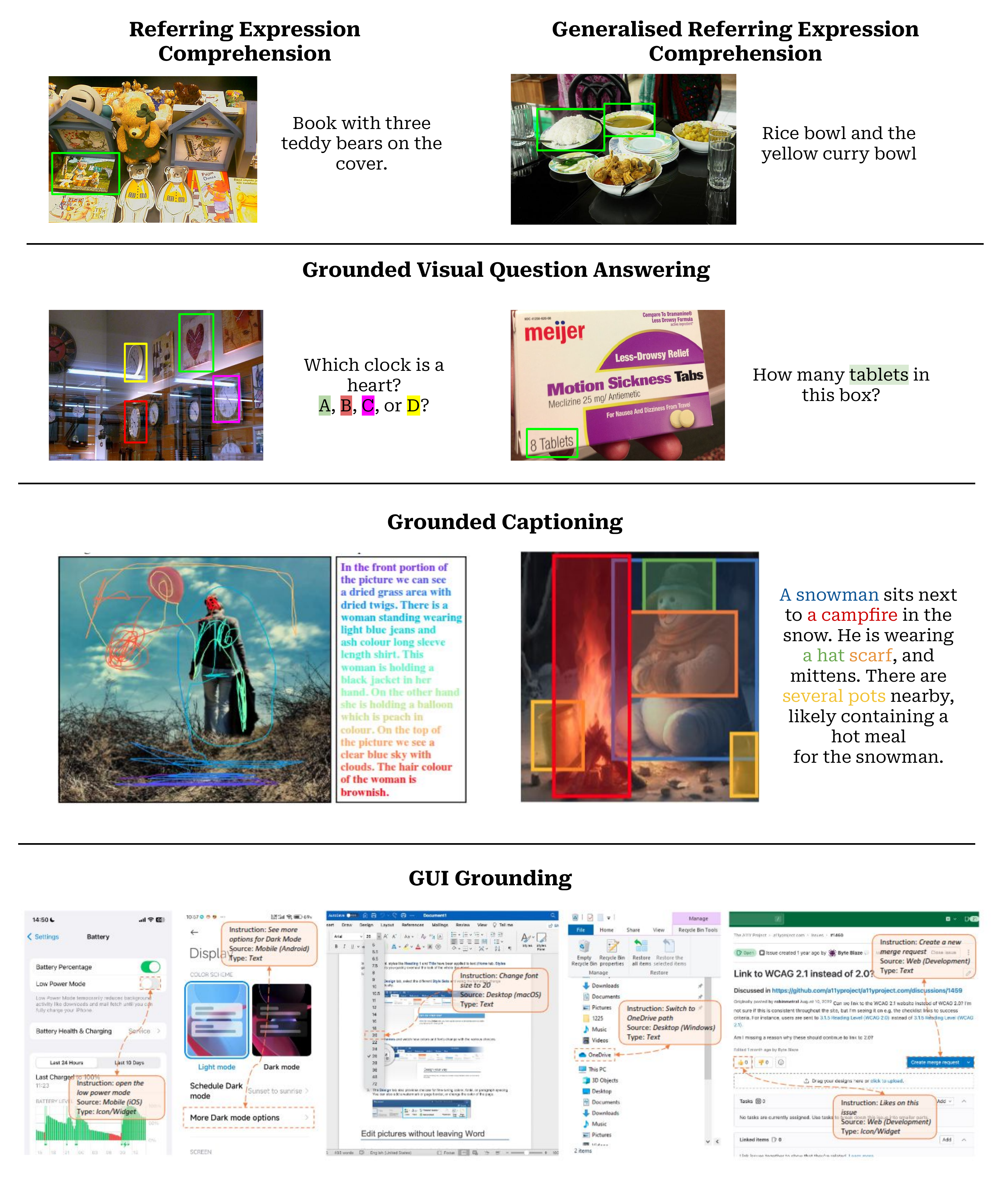}
    \caption{Inputs-output pairs from representative datasets across various domains related to visual grounding: 1) Referring Expression Comprehension (RefCOCOg \citep{mao2016generation}), 2) Generalised Referring Expression Comprehension (gRefCOCO \citep{he2023grec}), 2) Grounded Visual Question Answering (Visual7W \citep{zhu2016visual7w}, VizWiz-VQA \citep{gurari2018vizwiz}), Grounded Captioning (Localized Narratives \citep{PontTuset_eccv2020}, GRIT \citep{peng2023kosmos}), and GUI Grounding (ScreenSpot \citep{cheng2024seeclick}). All sources are listed from left-to-right in each respective domain.}
    \label{fig:examples_fig}
\end{figure*}

\subsection{Research Domains}
This section highlights key research domains where visual grounding is prevalent and outlines directions for  more in-depth analysis in the subsequent sections. \cref{fig:examples_fig} illustrates input-output pairs from representative datasets across research domains related to visual grounding.

\paragraph{Referring Expression Comprehension (REC)} REC \citep{kazemzadeh-etal-2014-referitgame, yu2016modeling, nagaraja2016modeling} requires the model to localise a specific region (i.e. a bounding box) within an image based on a given textual description. Similarly, Referring Expression Segmentation (RES) \citep{lai2024lisa} tasks the model with identifying an image-level mask that satisfies the query. Both
REC and RES rely on the strong assumption that a sentence describes exactly one object within an image, an assumption that often does not hold in real-world scenarios. Consequently, earlier models struggled when handling expressions referring to multiple or novel visual entities. More recently, Generalised Referring Expression Comprehension (GREC) \citep{he2023grec} and Generalised Referring Expression Segmentation (GRES) \citep{liu2023gres} have been proposed, which involve grounding on one, multiple, or even no objects described by the textual input within an image.

\paragraph{Grounded Visual Question Answering (GVQA)} GVQA requires the model to link answers to questions based on fine-grained visual observations. In this setting, visual grounding is required at the answer-level, where the model either provides a region in the image corresponding to the answer of the input question \citep{zhu2016visual7w, chen2022grounding, gan2017vqs}, or outputs both a textual answer and the associated region \citep{vogel2025refchartqa}. Grounding may also be necessary for interpreting the question itself, particularly when the visual prompt refers to or points at specific objects in the scene \citep{mani2020point, deitke2024molmo,Aho:72}.

\paragraph{Grounded Captioning (GC)} Grounded captioning refers a model's ability to generate captions that are explicitly anchored in visual regions. Existing work often produces multiple region-level descriptions \citep{krishna2017visual}, a setting commonly referred to as \textit{dense captioning}. In contrast, other approaches generate more holistic captions that describe the entire image while embedding references to specific regions directly within the description \citep{peng2023kosmos, gonzalez2021panoptic}. Some methods also produce visual traces that align words in the captions with corresponding regions in the image \citep{PontTuset_eccv2020}. More recently, grounded captioning has been extended to conversational settings \citep{ma2024groma}, in which the model generates grounded responses within a dialogue.

\paragraph{Agents Interacting with Graphical User Interfaces} Grounding plays a crucial role in enabling interactive agents to perceive, reason about, and act within their environment.
In this context, the textual query is often formulated as an instruction (e.g. execute a database search for specific records, or update a spreadsheet with newly provided values) rather than a description of an object in the scene. The agent then predicts the correct sequence of actions that satisfies the instruction. Thus, grounding not only allows the agent to interact effectively with elements in the GUI \cite{nguyen2025guiagentssurvey}, but also plays a critical role in task success, as a single grounding error can cause the agent to stall and fail to complete the task.

\paragraph{Grounded Reasoning} Inspired by the recent advances in LLMs \citep{guo2025deepseek, shao2024deepseekmath}, several approaches have adopted reinforcement learning to enhance multimodal reasoning capabilities by designing rule-based rewards tailored to specific downstream tasks \citep{kim2025robot, wang2025pixelthink}. Since grounding provides a flexible medium with interpretable explanations,  two promising avenues have emerged for grounded VLMs. First, grounding can be viewed as an intermediate reasoning step, where rewards are designed to align with the model's reasoning trace \citep{fan2025grit}. Second, the model can be directly rewarded based on its grounding performance at the solution level, without imposing constraints on the intermediate reasoning process \citep{shen2025vlm}. This remains an active area of research with the potential to unlock new capabilities of modern VLMs across the previously discussed domains.

\section{Developing Grounding Visual Language Models}
\label{sec:grounding_vlms}

This section provides an overview of the development of VLMs, with a focus on architectural design choices and training regimes that influence grounding performance.  The overview first outlines general design and training components that affect downstream applications, with particular emphasis on grounding capabilities. 

\begin{figure*}[h]
    \centering
    \includegraphics[width=\linewidth]{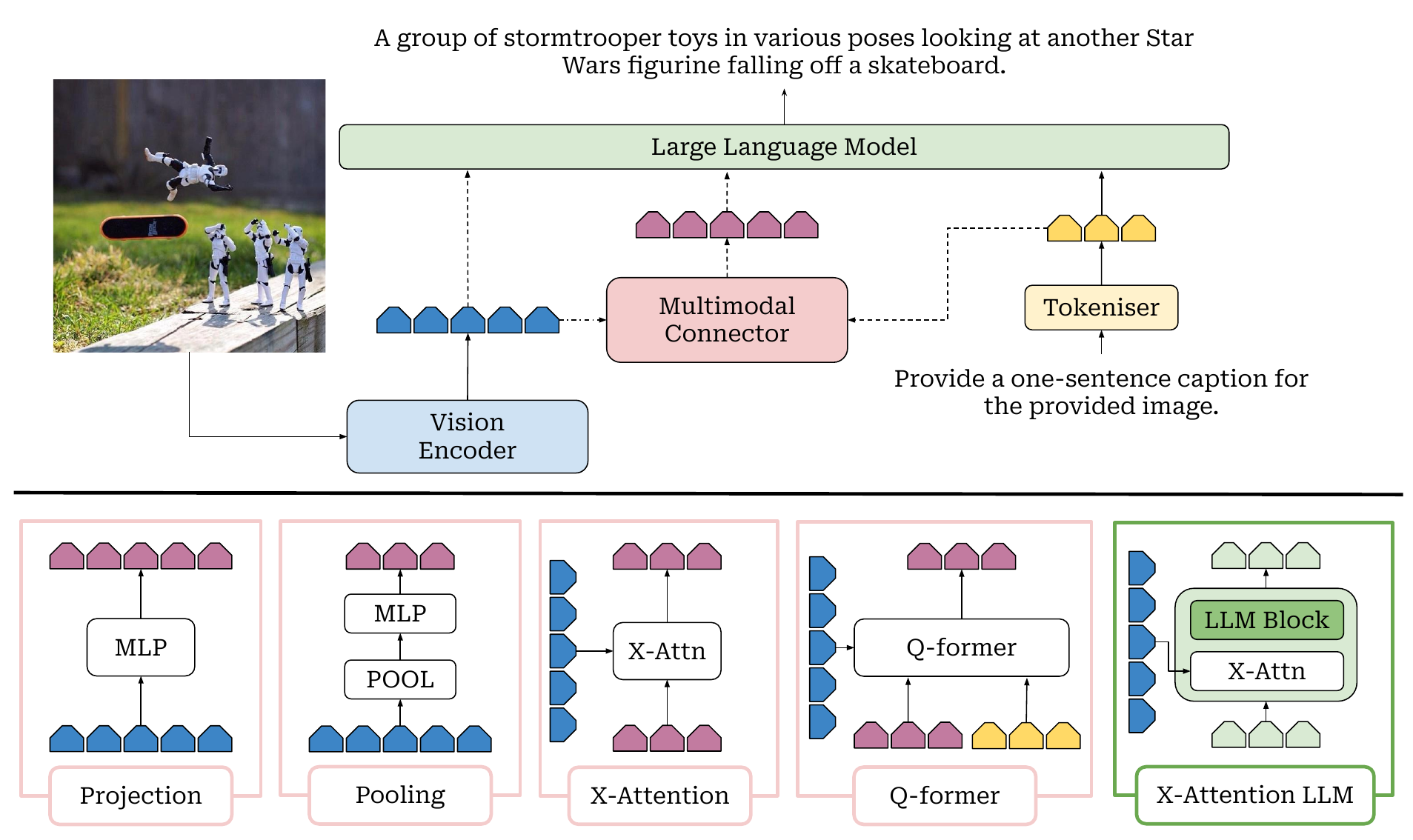}
    \caption{Architectural design choices in modern VLMs determine how different modalities are integrated. Typically, the language model accepts a sequence of image embeddings along with the textual input. These embeddings can be pre-processed by a multimodal connector that either preserves or compresses their representations. While most methods adopt an early fusion strategy before the language backbone \citep{bai2025qwen2, liu2023improved}, some approaches interleave modality-fusion layers directly in the language model \citep{alayrac2022flamingo, grattafiori2024llama}.}
    \label{fig:model_architectures}
\end{figure*}

Early work demonstrated the potential of combining Large Language Models (LLMs) with pre-trained vision encoders across diverse multimodal tasks \citep{alayrac2022flamingo, tsimpoukelli2021multimodal}. These studies motivated a paradigm shift away from complex multimodal architectures, explicit modality fusion, and specialised objectives \citep{chen2020uniter, tan2019lxmert, lu2019vilbert}, towards a simplified formulation, in which patch-based representations from a vision encoder are treated as token embeddings for input to a language model. \cref{fig:model_architectures} illustrates the architectural design of modern VLMs.

Most contemporary models \citep{bai2023qwen, bai2025qwen2, yang2025qwen3, laurenccon2024obelics, laurenccon2024matters, liu2024visual, chen2023pali} follow a common template comprising three components: a visual encoder \citep{zhai2023sigmoid, radford2021learning, fang2023eva, groeneveld-etal-2024-olmo}, a connector module, and a language backbone \citep{touvron2023llama, jiang2023mistral, team2024gemma}. The connector aligns the visual and textual embedding spaces, typically through a linear projection, typically through a pooling mechanism to downsample the image sequence, or more advanced designs that incorporate standalone or interleaved transformer blocks within the language model. These architectural choices are discussed in more detail below. \cref{tab:grounding_vlms} provides an overview of VLMs equipped with grounding capabilities. 

\begin{table*}[h]
    \centering
    % \scriptsize
    \small
    \renewcommand{\arraystretch}{1.2}
    % \begin{adjustbox}{center}
    % \addtolength{\tabcolsep}{-0.5em}
        \begin{tabular}{@{}l c c c c l@{}} 
        \textbf{Model} & \textbf{LLM} & \textbf{LLM type} & \textbf{Vision Encoder} & \textbf{Connector} & \textbf{Grounding Capabilities} \\  
        \toprule
        Mamba-VL \citep{pantazopoulos-etal-2024-shaking} & Mamba & M & EVA & MLP & REC, GC, RC, GVQA\\
        BuboGPT \citep{zhao2023bubogpt} & Vicuna & T & EVA & Q-former & GC\\
        GPT4RoI \citep{zhang2025gpt4roi} & Vicuna & T & CLIP & MLP & REC, GC\\
        Shikra \citep{chen2023shikra} & Vicuna & T & CLIP & MLP & REC, GVQA\\
        Kosmos-2 \citep{peng2023kosmos} & Kosmos-1 & T & BEiT & X-Attn & REC, RC\\
        GLaMM \citep{rasheed2024glamm} & Vicuna & T & OpenCLIP & MLP & REC, RC, GC\\
        NExT-Chat \citep{zhang24next} & Vicuna & T & CLIP & MLP & REC, RC, GC\\
        VistaLLM \citep{Pramanick_2024_CVPR} & Vicuna & T & EVA & Q-former & REC, GREC \\
        VisionLLM \citep{wang2023visionllm} & Alpaca & T & Intern & Q-former & REC\\
        MiniGPT-v2 \citep{chen2023minigpt} & Vicuna & T & CLIP & MLP & REC\\
        LLaVA-G \citep{zhang2024llava} & Vicuna & T & CLIP & MLP & REC\\ 
        G-GPT \citep{li-etal-2024-groundinggpt} & Vicuna & T &  CLIP & MLP & REC\\
        Groma \citep{ma2024groma} & Vicuna & T & CLIP & MLP & REC, RC, GC\\
        Qwen-VL \citep{bai2023qwenvlversatilevisionlanguagemodel} & Qwen & T & OpenCLIP & Q-former & REC, GC, RC\\
        VisCoT \citep{shao2024visual} & Vicuna & T & CLIP & MLP & REC, GVQA\\
        u-LLaVA \citep{xu2024u} & Vicuna & T & CLIP & MLP & REC\\
        CogVLM \citep{wang2024cogvlm} & Vicuna & T & EVA-2 & MLP & REC, GVQA\\
        Ferret \citep{youferret} & Vicuna  & T & CLIP & SA-Resampler & REC, GC, RC\\
        Ferret-v2 \citep{zhang2024ferret} & Vicuna & T & CLIP & SA-Resampler & REC\\
        ContextDET \citep{zang2025contextual} & OPT & T & Swin-Trans & MLP \& AvgPool & REC\\
        Internl-VL 2.5 \citep{chen2024expanding} & InternLM2.5 & T  & CLIP & MLP & REC\\
        Molmo \citep{deitke2024molmo} & Qwen2 & T & CLIP & MLP & GVQA\\
        Qwen-VL-2.5 \citep{bai2025qwen2} & Qwen2.5 & T & ViT & MLP$_s$ & REC\\
        Uground \citep{gou2025navigating} & Qwen2 & T & ViT & MLP$_s$ & AG\\
        SeeClick \citep{cheng2024seeclick} & Qwen & T & OpenCLIP & Q-former & AG\\
        CogAgent \citep{hong2024cogagent} & Vicuna & T &  EVA2 & X-Attn & AG\\
        Aria-UI \citep{yang2024aria} & Aria & T & SigLIP & MLP & AG\\
        ShowUI \citep{lin2025showui} & Qwen2 & T & CLIP & MLP & AG\\
        OS-Atlas \citep{wu2024atlas} &  Qwen2 & T & CLIP & MLP & AG\\
        Kimi-VL \citep{team2025kimi} & Moonlight & T & SigLIP & MLP & AG\\
        Jedi \citep{xie2025scaling} & Qwen 2.5 & T &  ViT & MLP$_s$ & AG\\
        Seed1.5-VL \citep{guo2025seed1} & Seed1.5-LLM & T & ViT & MLP$_s$ & REC, GVQA, AG\\
        \bottomrule
        \end{tabular}
    
    \caption{Summary of grounding VLMs. (i) Grounding Capabilities: Referring Expression Comprehension (REC), Generalised Referring Expression Comprehension (GREC), Grounded Captioning (GC), Region Captioning (RC), Grounded Visual Question Answering (GVQA), Agent Grounding (AG); (ii) LLM type: Transformer-based (T), Mamba-based language backbone (M); and (iii) Multimodal Connector: Multilayer perceptron (MLP) matching the dimensionalities of visual and textual embeddings with a projection block. (MLP$_s$): merging neighbor sets of patch features before applying the MLP block. (Q-former) a transformer stack with learnable queries that cross-attend at the visual embeddings, downsampling the input visual sequence. (SA-Resampler): using point clouds from image regions to downsample visual tokens. (X-Attn) a single cross attention layer with learnable queries similar to Q-former.}
    \label{tab:grounding_vlms}
\end{table*}

The vision encoder provides the language model with visual representations. Most approaches adopt CLIP-based architectures \citep{ radford2021learning, fang2023eva, wortsman2022robust}, whose pretraining objective is the alignment between textual and visual embeddings. The language backbone is generally a transformer-based language model, although some studies \citep{pantazopoulos-etal-2024-shaking, zhao2024cobra, qiao2024vl} have explored alternatives such as Mamba \citep{gu2023mamba}, a structured state-space model that matches or surpasses transformer performance in sequence modelling while offering improved inference efficiency and linear scaling with sequence length. To the best of our knowledge, grounding with a Mamba backbone has been explored in a limited number of studies; the findings in \citep{pantazopoulos-etal-2024-shaking} indicate that transformer-based backbones perform significantly better on grounding tasks.

In addition, a connector module bridges the two modalities by matching their embedding dimensions. Most recent models adopt a simple projection mechanism \citep{chen2023shikra, bai2025qwen2, li2024llava}. A common challenge arises because the number of visual patch embeddings often exceeds the number of textual tokens. This imbalance can increase training and inference times, especially in applications involving high-resolution images, videos, or large multimodal documents. One solution is to compress the visual sequence using a resampler component, which may take the form of average pooling \citep{zang2025contextual}, attention mechanisms before the language model\citep{peng2023kosmos}, or multi-stage processes (e.g. \citep{laurenccon2024building}). Other designs \citep{alayrac2022flamingo, dai2024instructblip, zhao2023bubogpt} interleave cross-attention modules within the language model itself.

%\begin{figure}[h]
%    \centering
%    \includegraphics[width=\linewidth]%{figures/alan_turing_training_larger.pdf}
%    \caption{The sequential stages of training and the types of data used throughout the development of a VLM \citep{laurenccon2024building}.}
%    \label{fig:training_stages}
%\end{figure}

Training VLMs is also a complex, resource-intensive task, with many open research questions regarding the contribution of individual components to downstream performance. Challenges related to reproducibility, proprietary weights, and inconsistent evaluation protocols complicate direct comparisons across models. Some studies \citep{fan2025grit, shen2025vlm, luo2025gui} have further explored reinforcement learning as an additional training phase, motivated by advances in language modelling \citep{guo2025deepseek, shao2024deepseekmath}.

In summary, the design, development, and training of VLMs constitute an intricate process. Recent work \citep{pantazopoulos-etal-2024-shaking, laurenccon2024matters, karamcheti2024prismatic} has analysed the effects of individual components in the VLM architecture. The following sections examine these components in greater detail and assess their impact on grounding capabilities.

\subsection{Vision Encoder} 
The most commonly employed visual encoders are based on pre-trained ViT models with a CLIP-based objective, which leverages the inherent alignment of CLIP embeddings \citep{radford2021learning, fang2023eva}. The standard image-text contrastive objective aligns image and text embeddings for matching (positive) pairs while ensuring that unrelated (negative) pairs remain dissimilar in the embedding space. This is achieved through a batch-level, softmax-based contrastive loss applied twice to normalise pairwise image-text and text-image similarity scores.

Although effective, this approach has two notable limitations. First, the computation requires two separate steps to estimate image-to-text and text-to-image similarities. Second, the degree of alignment depends on the batch size, since larger batches can provide more negative pairs. These limitations are addressed in SigLIP \citep{zhai2023sigmoid, tschannen2025siglip}, where the sigmoid loss eliminates the need for operations across the full batch, thereby reducing the requirement to distribute the loss and improving efficiency. Consequently, many subsequent VLMs \citep{laurenccon2024matters, marafioti2025smolvlm, laurenccon2024building} adopt the SigLIP variation, although these models are not specifically designed for multimodal grounding applications. A side-by-side comparison for REC is presented in \citep{karamcheti2024prismatic}, showing that SigLIP slightly outperforms CLIP.

\subsubsection{Image Resolution}

Recent research has demonstrated that providing high-resolution images to VLMs reduces hallucinations and enhances multimodal understanding, leading to improved performance on tasks requiring fine-grained details \citep{wang2022ofa, bai2025qwen2, pantazopoulos-etal-2024-shaking, liu2023improved, li2024llava, karamcheti2024prismatic, zhang24next, liu2024llavanext, chenpali, beyer2024paligemma, steiner2024paligemma, mckinzie2024mm1}.
However, most off-the-shelf vision encoders operate on low-resolution images and do not enforce a fixed square aspect ratio. This creates practical challenges: (i) models trained on smaller images must adapt to larger ones at inference; and (ii) the increased number of tokens leads to higher computational costs in both the visual encoder and the language model.
To address these issues, two broad categories of approaches have been proposed, aimed at ensuring that the language model receives high-quality visual representations.

\paragraph{Positional-Encoding Interpolation} A common method for handling high-resolution images is to interpolate the positional encodings of the visual encoder. This approach has been employed in earlier multimodal models trained in a multi-task fashion \citep{wang2022ofa, yang2022unitab, li2021align} and has since been adopted by many VLMs \citep{beyer2024paligemma, steiner2024paligemma, arora2023zoology} which require extensive training with high-resolution images during the later stages of development. In addition, several studies propose reducing the visual sequence by merging neighbouring tokens \citep{bai2023qwen, chen2023minigpt}. Visual token compression is discussed in \cref{sec:mm_connector}, where it is treated as a function of the multimodal connector component.

\paragraph{Handling Arbitrary Resolution Images}
Certain approaches address high-resolution inputs by dividing the image into fixed-size sub-images that match the expected dimensions of the visual encoder \citep{bai2025qwen2, liu2024llavanext, liu2024sphinx, li2024monkey, guo2024llava}. Each sub-image is processed independently by the visual backbone, while the original image is downscaled to the same resolution and processed in parallel. The features extracted from all sub-images, together with those from the downscaled whole image, are then concatenated to form a global representation.

\subsection{Connecting Visual and Textual Representations in VLMs}\label{sec:mm_connector}

Two principal strategies have emerged for connecting visual and textual embeddings: directly aligning the dimensionality of the two modalities, and compressing the visual sequence. The former, described in \citep{mckinzie2024mm1, lin2024preserve}, is regarded as a \textit{feature-preserving} method because it maintains the number of feature maps. In contrast, the latter is considered as a \textit{feature-compressing} method, as it incorporates a (learnable) component that reduces the number of patch embeddings.

\subsubsection{Mapping Visual to Textual Representations}
Since the LLM backbones are primarily trained on generic text, an inherent semantic gap arises when processing multi-modal features. Mapping operations 
address this gap by transforming the visual embeddings into the dimensionality expected by the language model. This transformation can be implemented as a simple linear projection or a stack of multilayer perceptrons (MLPs).
The resulting embeddings are then concatenated with the text input sequence without any compression.

\paragraph{Linear Transformation} The most basic transformation of visual embeddings employs a single linear layer, which was first introduced in \citep{liu2024visual, pmlr-v202-koh23a}. However, few subsequent models have adopted this approach \citep{ye2024mplug}, partially because non-linear transformations provide greater representational capacity.

\paragraph{Non-Linear Transformation using MLPs} A natural extension introduces non-linearity into the feature transformation process by employing an MLP. This approach consistently enhances the model's capabilities across a wide range of multimodal tasks \citep{liu2023improved}, while ensuring that the full visual information is preserved, as no pooling strategy is applied.
Many subsequent VLMs have adopted this idea \citep{bai2025qwen2, chen2024expanding, li2024llava, marafioti2025smolvlm, liu2024llavanext}, as it offers a favourable balance between simplicity of design and performance.

\subsubsection{Compressing the Input Sequence}
Despite their conceptual simplicity, mapping methods produce long sequences of visual tokens, which reduce the efficiency of both training and inference. 
Based on prior works we identify four approaches aimed at compressing the input sequence.

\paragraph{Pooling} Pooling is among the most basic approaches for compressing visual embeddings, whereby adjacent patch embeddings are merged through a pooling operation and subsequently projected to match the dimensionality of the language embeddings. Two VLMs \citep{bai2025qwen2, chen2023minigpt} that adopt this strategy combine four spatially adjacent patch features into a single vector through an MLP before passing them to the large language model.

\paragraph{Convolution} CNNs offer a distinct advantage for multimodal learning by preserving local spatial structures and enhancing spatial understanding through their hierarchical feature extraction capabilities. Unlike traditional pooling methods, which may discard important spatial relationships, CNNs retain geometric and positional information that is critical for capturing fine-grained spatial details, particularly in multimodal grounding tasks. Two notable approaches exemplify this idea: C-Abstractor \citep{cha2024honeybee}, which reintroduces two-dimensional (2D) positional embeddings into the visual features, and is subsequently followed by ResNet block \citep{xie2017aggregated}, and the H-Reducer \citep{hu2024mplug}, which employs CNNs to reduce the number of image hidden states by a factor of four.

\paragraph{Cross-Attention Module} An alternative approach to compression involves introducing learnable embeddings that cross-attend to the image representations. Since the number of patch embeddings generally exceeds the number of learnable embeddings in the module, this mechanism naturally results in compression. Qwen-VL \citep{bai2023qwenvlversatilevisionlanguagemodel} adopts this strategy by reducing the number of visual tokens through a single-layer cross-attention module operating between a set of learnable embeddings and the image hidden states.

\paragraph{(Multimodal) Resamplers} Similar to the cross-attention module, a resampler component is implemented as a transformer comprising stacked self- and/or cross-attention modules and learnable queries. We identify two notable resamplers proposed in prior work: the Perceiver \citep{pmlr-v139-jaegle21a} and the Q-former \citep{li2023blip}. The Perceiver has been applied alongside interleaved cross-attention layers on the language model \citep{alayrac2022flamingo, laurenccon2024matters, awadalla2023openflamingo, li2025otter}, wherein the language tokens interact with the compressed visual sequence only through these interleaved cross-attention layers. The Q-former has likewise been adopted in subsequent studies\citep{dai2024instructblip, hu2024bliva,  chen2024lion}, demonstrating its effectiveness in various settings.

Despite their efficiency, both resamplers have been critiqued in several studies \citep{lin2024preserve, pantazopoulos-etal-2024-lost}, which argue that these components do not necessarily preserve spatial relationships but instead compress the visual embeddings into a bag-of-words-like representation. While feature-preserving approaches offer superior performance, their advantage over feature-compressing connectors diminishes rapidly as image resolution increases. 

Moreover, connectors based on CNNs \citep{cha2024honeybee} consistently outperform attention-based resamplers across all resolutions and task granularities, as they more effectively exploit the two-dimensional structure of images. Finally, some models \citep{youferret, zhang2024ferret} adopt alternative resampling strategies, such as point-clouds, image segmentation, and clustering. Although these approaches have shown promising results, there is currently little evidence supported by rigorous and fair quantitative evaluation.

\subsection{Language Model}

\paragraph{Language Backbones Alternative to Transformers} The vast majority of modern VLMs employ a transformer-based language backbone. Although a few studies have explored Mamba-based models \citep{zhao2024cobra, qiao2024vl} and reported promising results, these findings do not conclusively demonstrate the effectiveness of a VLM with a Mamba-based language backbone, as the compared models were of similar size but trained under substantially different regimes.

The analysis in \citep{pantazopoulos-etal-2024-shaking} provides a more detailed examination. However, the models considered are relatively small and not instruction-tuned. Moreover, while earlier work has shown that the quality of language model directly correlates with the overall performance of the VLM \citep{laurenccon2024matters}, the findings in \citep{pantazopoulos-etal-2024-shaking} indicate that this relationship depends on the downstream task. For example, in grounding and generation tasks, the transformer-based models consistently outperformed their Mamba-based counterparts across all model scales.

\paragraph{Language Models without Vision Encoders} Recent approaches have explored the use of language backbones without explicitly generating visual representations through a dedicated vision encoder. The Fuyu model \citep{fuyu-8b} exemplifies this paradigm by bypassing traditional vision encoders and instead feeding image patches directly into the language model via a simple linear projection to adjust their dimensionality. This architecture offers two key advantages: it operates independently of any pre-trained vision model, and it preserves the complete information from the original image, which can be critical for accurate prompt responses. 

However, despite these theoretical benefits, the direct patch approach has not demonstrated superior performance in practice. Fuyu performs significantly worse than similarly-sized contemporary models on standard benchmarks, and PaliGemma \citep{beyer2024paligemma}. Empirical evaluations report substantial performance degradation compared to models that incorporate pre-trained vision encoders. Moreover, processing image representations directly within the language model may impair its effectiveness on text-only tasks, although this hypothesis remains largely untested, as most VLMs are not evaluated on pure textual benchmarks. Finally, the absence of efficient pooling strategies for managing raw pixel data poses scalability challenges in applications that require processing large numbers of visual tokens.

\subsection{Training Pipeline}
The development of a VLM typically begins with unimodal pre-trained backbones, consisting of a language model and a vision encoder. Some earlier works, however, adopt a single-stage training procedure \citep{alayrac2022flamingo, tsimpoukelli2021multimodal, pmlr-v202-koh23a, awadalla2023openflamingo, li2025otter, gao2023llama}. 

Most contemporary approaches employ a more complex pipeline involving multiple stages \citep{bai2025qwen2, chen2024expanding, liu2023improved, laurenccon2024matters, liu2024llavanext} (see \citep{laurenccon2024building} for a detailed discussion of training stages). †t
This design is motivated primarily by the limited availability of high-quality data at scale, memory constraints that hinder efficient training, and stability considerations \citep{laurenccon2024building}. During these stages, progressively higher-quality data is introduced, the maximum image resolution is gradually increased, and additional components of the model are unfrozen.

\subsubsection{Image-text alignment pretraining} 

The primary goal of pre-training is to align the backbone models and to train the newly initialised parameters. During this stage, the model is exposed to examples that support the following capabilities: image captioning, handling an arbitrary number of images interleaved with diverse texts, reading comprehension, and grounding capabilities, all of which are essential for downstream applications. Training usually leverages image-text pairs or large-scale web-derived image-text datasets, which are generally constructed by crawling the web, downloading images, and extracting the corresponding textual descriptions from the original HTML files \citep{schuhmann2022laion, kakaobrain2022coyo-700m, gadre2023datacomp}.

Owing to the relative ease of collection, such raw image-text pairs are widely used; however, their effectiveness in establishing strong alignment between images and text varies considerably, as the accompanying alt-texts are often noisy, ungrammatical, or overly brief \citep{gadre2023datacomp}. These issues can introduce challenges during training. Recent approaches have sought to mitigate this by applying synthetic re-captioning \citep{mckinzie2024mm1, li2023blip, shi2020improving, lai2024veclip}, and quality-filtering techniques \citep{sharifzadeh2024synth, abbas2023semdedup}, both of which have demonstrated improved performance.

Training on interleaved image-text documents was first introduced in \citep{alayrac2022flamingo, laurenccon2024obelics}, as it enhances: (i) in-context learning abilities; (ii) the model’s ability to process an arbitrary number of images interleaved with text; and (iii) the model's exposure to  a wider distribution of content beyond standard image-text pairs.
These claims have been validated by other works \citep{mckinzie2024mm1} and subsequently adopted in follow-up models \citep{bai2025qwen2, chen2024expanding, ye2024mplug, li2024llava, zhang2024internlm, xue2024xgen}. 

For reading comprehension tasks, it is common to train on PDF-text pairs \citep{biten2022ocr} or to employ OCR tools to extract text from images \citep{zhang2023pmc, yao2024minicpm, textocr-gpt4v}. Finally, datasets supporting grounding capabilities require a more complex pipeline, which generates region-level annotations using segmentation, detection, or tagging models. 
Two notable works in this direction are GRIT \citep{peng2023kosmos} and BLIP3-GROUNDING-50M \citep{xue2024xgen}. 

GRIT is a collection of image-text pairs constructed via a pipeline that extracts and links text spans, noun phrases, and referring expressions in captions to corresponding image regions. In the first step, the pipeline extracts noun chunks from captions using spaCy \citep{spacy2} and associates these chunks with bounding boxes via GLIP \citep{Li_2022_CVPR}, a pre-trained grounding model. In the second step, spaCy is used to identify dependency relations in the caption, and the noun chunks are expanded into referring expressions by recursively traversing nodes in the dependency tree.
Similarly, in BLIP3-GROUNDING-50M, captions for the associated objects and their locations are obtained from state-of-the-art image tagging \citep{Zhang_2024_CVPR} and detection \citep{liu2024grounding} models.

\subsubsection{Fine-tuning}
Fine-tuning VLMs is largely influenced by the training paradigm of LLMs and generally comprises two stages: supervised fine-tuning (SFT) followed by an alignment phase. The former has been well established and constitutes a standard component in the development of most VLMs, whereas the latter remains underexplored, particularly in the context of grounded VLMs.

\paragraph{Teaching VLMs to Follow Instructions} The concept of enabling VLMs to respond to natural language instructions is heavily inspired by earlier work on instruction-following LLMs \cite{ouyang2022training, weifinetuned, sanhmultitask} and was first extended to VLMs in later works \citep{dai2024instructblip, li2023seed}. At this stage, various datasets, such as those described in \cref{sec:approaches}, are reformulated into instruction-response pairs, often adopting a standardised format that supports fine-tuning. These instruction-tuning datasets commonly encompass a wide range of tasks, including image captioning, visual reasoning, and grounding, thereby aligning the outputs of VLMs more closely with human-annotated reference outputs. Standardisation of the instruction format is crucial because it allows the model to generalise to unseen instructions and to interact effectively in multimodal scenarios.

\paragraph{Improving Alignment and Multimodal Reasoning via Reinforcement Learning} The final stage in the development cycle of a VLM is the alignment stage, in which the model learns directly from feedback on its outputs. This phase aims to reduce hallucinations and the risk of generating harmful or inappropriate responses, while also potentially enhancing overall performance \citep{yao2024minicpm, zhang2025spa}. An additional benefit of this phase is the improvement of the model's multimodal reasoning capabilities.

A key observation in LLMs is that chain-of-thought (CoT) reasoning traces, as commonly employed in natural language processing, often lack an effective format for certain tasks, particularly in multimodal settings. Although CoT traces decompose complex problems into step-by-step natural language explanations, making the output more intuitive and accessible to humans, much of the generated linguistic content can be regarded as superfluous from the model's perspective, as it does not directly contribute to task success. Motivated by this, recent work has proposed Grounded Policy Refinement Optimisation (GPRO) \citep{guo2025deepseek, shao2024deepseekmath}, a reinforcement learning framework in which the model generates outputs without being constrained by teacher-forced CoT solutions. In GPRO, the model receives a reward based on a combination of signals, such as the correctness and formatting of the solution, as well as task-specific objectives tailored to the downstream application. This rule-based reward design is flexible and can be adapted to various multimodal tasks, for example, by directly optimising evaluation metrics relevant to grounded generation (see \cref{sec:approaches}).

Contemporary research explores this direction from two complementary perspectives: (i) grounding as a reasoning process; and (ii) reasoning in service of grounding in VLMs. In the first line of work, grounding signals are integrated into the CoT trace \citep{fan2025grit} itself, resulting in explanations that are both interpretable and grounded in the input modalities. In the second approach, the model is rewarded for its grounding capabilities at the level of the final solution, without imposing constraints on the intermediate reasoning trace \citep{shen2025vlm}.

Although both approaches have demonstrated promising results, further research is needed to fully realise their potential and to establish best practices for alignment and multimodal reasoning in VLMs.

\begin{table*}[ht!]
    \centering
    \scriptsize
    \renewcommand{\arraystretch}{1.2}
    % \begin{adjustbox}{center}
    \addtolength{\tabcolsep}{-0.4em}
        \begin{tabular}{@{}l c c c c p{8.5cm}} 
        \textbf{Dataset} & \textbf{Domain} & \textbf{Train} & \textbf{Eval} & \textbf{Annotation} & \textbf{Short Description}\\  
        \midrule
        RefCOCO \citep{yu2016modeling} & REC & \cmark & \cmark & H & Referring expressions collected via a collaborative describer-guesser game on MSCOCO images and categories.\\
        RefCOCO+ \citep{yu2016modeling} & REC & \cmark & \cmark & H & Referring expressions collected via a collaborative describer-guesser game on MSCOCO images and categories.\\
        RefCOCOg \citep{mao2016generation} & REC & \cmark & \cmark & H & Referring expressions collected via a collaborative describer-guesser game on MSCOCO images and categories.\\
        Ref-L4 \citep{chen2025revisiting} & REC & \cmark & \cmark & S + H & Repurposed examples from RefCOCO/+/g sources with longer expressions and larger vocabulary.\\
        gRefCOCO \citep{he2023grec} & GREC/GRES  & \cmark & \cmark & H & Repurposed annotations from RefCOCO, where the referring expression can match one, multiple, or no target recipients.\\
        LISA \citep{lai2024lisa} & GRES & \cmark & \cmark & S + H & Instructions where the expected response is a set of segmentation masks.\\
        \midrule
        Visual-7W \citep{zhu2016visual7w} & GVQA & \cmark & \cmark & H & Multiple-choice questions where each choice depicts a candidate bounding box in the image.\\
        PointQA \citep{mani2020point} & GVQA & \cmark & \cmark & H & Questions refer to singular pixels in images, under three settings: 1) the region around the point is relevant to the question,
2) global understanding, as the region points to an element that may occur multiple times in the image,
and 3) a general setting by repurposing examples from Visual7W.\\
        VizWiz-VQA \citep{chen2023vqa} & GVQA & \cmark & \cmark & H & Includes images taken from BLVs and the task is to return the region in the image used to arrive at the answer to the visual question.\\
        RefChartQA \citep{vogel2025refchartqa} & GVQA & \cmark & \cmark & H & Answering questions related to chart images by providing the corresponding region.\\
        VCR \citep{zellers2019recognition} & GVQA & \cmark & \cmark & H & Multiple-choice questions where each choice describes an explanation linked to entities in the image.\\
        Visual Coment \citep{park2020visualcomet} & GVA & \cmark & \cmark & H & Generate a set of commonsense inferences on events before, at present, and after the visual context\\
        Sherlock \citep{hessel2022abduction} & GVA & \cmark & \cmark & H & Given localised visual clues the objective is to provide commonsense inferences about events in images\\
        \midrule
        Visual Genome \citep{krishna2017visual} & GC & \cmark & \xmark & H & Dense captions collected from humans describing regions in images.\\
        GRIT \citep{peng2023kosmos} & GC & \cmark & \xmark & S & Image-grounded caption pairs, where the images and text are crawled from the web. Captions grounded into the image using an object detection and dependency parsing pipeline.  \\
        blip3-grounding-50m \citep{xue2024xgen} & GC & \cmark & \xmark & S & Image-grounded caption pairs, where the images and text are crawled from the web. Captions grounded into the image using an image tagging and dependency parsing pipeline.  \\
        GroundedCap \citep{oliveira2025groundcap} & GC & \cmark & \xmark & S & Dense captions collected .\\
        Widget Caption \citep{li2020widget} & GC & \cmark & \cmark & H & Captions describing UI elements Image-grounded caption pairs, where the image are collected from movie scenes.\\
        Localized Narratives \citep{PontTuset_eccv2020} & GC & \cmark & \xmark & H & Grounded captions where annotators were tasked with describing the image while simultaneously hovering their mouse over the respective regions.\\
        \midrule
        GUIAct \citep{chen2024guicourse} & AG & \cmark & \cmark & S + H & Collection of three datasets for GUI understanding focusing on OCR and grounding abilities, GUI navigation, and conversations.\\
        ScreenSpot \citep{cheng2024seeclick} & AG & \xmark & \cmark & H & Expressions from various environments, including iOS, Android, macOS, Windows, and Web.\\
        ScreenSpot-v2 \citep{wu2024atlas} & AG & \xmark & \cmark & H & Repurposed annotations from ScreenSpot.\\
        ScreenSpot-Pro \citep{li2025screenspotpro} & AG & \xmark & \cmark & H & Instructions covering high-resolution images focusing on four types of professional applications.\\
        OSworld-G \citep{xie2025scaling} & AG & \xmark & \cmark & H & Expressions targeting GUI elements that cover text matching, element recognition, layout understanding, fine-grained manipulation and infeasibility.\\
        Aria \citep{yang2024aria} & AG & \cmark & \xmark & S & Collection of existing training resources for developing GUI agents for web and android, and desktop interfaces.\\
        Uground \citep{gou2025navigating} & AG & \cmark & \xmark & S & Collection of existing training resources for developing GUI agents for web and android.\\
        Jedi \citep{xie2025scaling} & AG & \cmark & \xmark & S & Collection of existing training resources for developing GUI agents with examples targeting icons, layouts, and GUI components.\\
        \bottomrule
        \end{tabular}

    \caption{Summary of existing datasets for developing and evaluating grounding VLMs. Train / Eval columns cells marked with \cmark or \xmark \enspace denote that the respective benchmark contains (or not) a subset for training or evaluation. Annotations are marked with (H) if the examples are labeled by humans or with (S) for synthetic examples.}
    \label{tab:datasets}
\end{table*}

\section{Evaluations}\label{sec:approaches}

This section presents the data resources pertinent to each evaluation domain. The accompanying table (see \cref{tab:datasets}) summarises the available benchmarks and datasets, providing a concise description of each. In addition, this section outlines the evaluation metrics commonly employed to assess the performance of VLMs across a variety of tasks. 

\subsection{Benchmarks and Datasets}

In order to evaluate the capabilities of grounded vision-language models, a variety of benchmarks and datasets have been introduced, providing the necessary tasks and data for training and evaluation. The main benchmarks and datasets are outlined as follows:

REC involves the task of localising a specific target instance in an image based on a given textual description. Research in this area is motivated in part by early work such as ReferItGame \citep{kazemzadeh-etal-2014-referitgame}, a two-player referring expression game.
In this setup, the first player is shown an image with a target object outlined in red and asked to compose a referring expression describing the object.
The second player sees the same image, along with the referring expression written by the first player, and clicks on the location of the described object. The interaction is deemed successful if the second player correctly identifies the target object corresponding to the expression.

The most widely used benchmarks for REC include RefCOCO \citep{yu2016modeling}, RefCOCO+ \citep{yu2016modeling}, and RefCOCOg \citep{mao2016generation}, developed by a similar two-player game using COCO images \citep{lin2014microsoft}.
ALthough existing grounding VLMs report strong performance \citep{chen2023shikra, bai2025qwen2, wang2024cogvlm, zhao2024llm} on these datasets, recent studies \citep{chen2025revisiting} have re-examined their suitability for evaluating REC capabilities of VLMs. Specifically, RefCOCO consists of relatively succinct expressions, while RefCOCO+ deliberately excludes locational terms to encourage the use of more semantically rich descriptions. RefCOCOg introduces more elaborate and descriptive annotations, arguably to address limitations in the earlier variants. Nevertheless, these benchmarks have significant shortcomings, including labelling errors such as typographical mistakes, misalignments between referring expressions and target instances, and inaccurate bounding boxes. As a result, some recent works have moved away from these established benchmarks. For example, 
Ref-L4 \citep{chen2025revisiting} is a corpus constructed from cleaned versions of the earlier datasets as well as additional images from Objects365 \citep{shao2019objects365}, offering a broader range of visual inputs compared to the original sources \citep{lin2014microsoft}. Finally, Ref-L4 extends the task further by introducing longer referring expressions and greater diversity in object categories and vocabulary.

Most REC/RES benchmarks assume that each sentence refers to exactly one object in the image. However, this assumption is often unrealistic in real-word scenarios.
To address this limitation, recent work has proposed GREC/GRES, where the input query may correspond to a single object, multiple objects, or no object at all.
gRefCOCO extends the existing REC benchmarks to accommodate both GREC \cite{he2023grec} and GRES \citep{liu2023gres} settings. 
Similarly, LISA \citep{lai2024lisa} approaches grounding from the GRES perspective, using images obtained from OpenImages \citep{kuznetsova2020open}, and ScanNetv2 \citep{dai2017scannet}. In this benchmark, however, the text query takes the form of a complex question about the image content that often requires background knowledge or common-sense reasoning about the objects depicted.

GVQA supports the development of models that provide visual evidence in a manner similar to how humans rely on visual cues to  answer questions. Only a few studies have focused on \textit{pointing} questions, where the answer refers directly to an entity in the image, as well as on  chart understanding, and commonsense reasoning. 

Visual-7W \citep{zhu2016visual7w} is one of the earliest works in this direction, including a pointing subset in which the model is provided with four candidate bounding boxes and selects the one corresponding to the correct answers. PointQA \citep{mani2020point} builds on Visual-7W, with questions more explicitly grounded in visual observations. Another line of work, VizWiz-VQA-Grounding \citep{chen2023vqa}, extends this idea by drawing inspiration from questions posed by blind or low-vision individuals \citep{gurari2018vizwiz}, addressing cases in which a visual question admits multiple valid natural language answers. Similarly, RefChartQA \citep{vogel2025refchartqa} targets visually-grounded answering for chart understanding, highlighting how grounding capabilities are closely tied to a model's ability to read textual content from images.

A separate line of research explores deriving commonsense inferences from grounded visual observations. In Visual Commensense Reasoning (VCR) \citep{zellers2019recognition}, the model is given an image, a list of regions, and multiple choice questions, and selects the most plausible answer along with an appropriate rationale. Subsequent work, VisualCOMET \citep{park2020visualcomet}, builds upon VCR by expanding the task to include inferring plausible events before and after the current image and identifying the intent underlying the present scene. Lastly, Sherlock \citep{hessel2022abduction} introduces a framework where the model is presented with a series of clues grounded in the image and scores a set of candidate inferences according to their plausibility.

Grounded captioning is commonly categorised into three main classes: (i) region captioning, which refers to approaches for describing regions in an image; (ii) holistic captions, where the entities in the description are explicitly linked to specific visual regions; and (iii) captions conditioned on both the image and the traces within it. 

Visual Genome \citep{krishna2017visual} is the most prominent representative of dense captioning, providing multiple region-level descriptions per image, with each caption associated with a specific region. However, similar to vanilla REC benchmarks, the dense captions in Visual Genome are often brief and exhibit limited linguistic variability.
Moreover, this approach restricts the ability to produce dense descriptions in which words can be simultaneously grounded to multiple objects and actions distributed across different regions of the image. The region-specific approach also constrains the capacity to capture scene descriptions that integrate both static elements and dynamic interactions within a single caption. Subsequent work in GC has shifted toward holistic captions, which describe the entire visual scene while grounding the entities to specific objects in the image. GRIT \citep{peng2023kosmos}, and BLIP3-GROUNDING-50M \citep{xue2024xgen} are two recent examples,  both compiling large collections of grounded image-text pairs from web-scrapped images \citep{gadre2023datacomp, schuhmann2022laion, kakaobrain2022coyo-700m}.
To construct aligned image-region descriptions, both approaches employ natural language processing tools \citep{spacy2}, image tagging models \citep{Zhang_2024_CVPR}, and detection models \citep{liu2024grounding} to extract noun phrases from collected captions and link these phrases to detected regions in the images. Notably, since these image-text pairs are automatically generated from web data, the captions are often noisy and provide only weak supervision signals; therefore, such datasets are primarily used for pre-training (see \cref{sec:grounding_vlms}). GroundedCap \citep{oliveira2025groundcap} attempts to mitigate some of these limitations by using images from movie scenes \citep{huang2020movienet}, although the majority of captions are still automatically generated through a similar pipeline. The final method, Localised Narratives \citep{PontTuset_eccv2020}, offers a more precise form of grounded captions, where annotators describe the image content aloud while simultaneously moving a mouse pointer over the corresponding regions. Because the speech and pointer trajectories are synchronised, every word in the description is explicitly linked to a pixel-level location, producing detailed trajectory traces that can be used to train models for grounded captioning.

A recent trend in artificial intelligence involves equipping agents with the ability to interact with the web \citep{deng2023mind2web}, operating systems \citep{xie2024osworld}, or mobile devices \citep{rawles2023androidinthewild}. Visual grounding is, therefore, a necessary prerequisite for such agents, as it enables them to identify and interact with key elements in a GUI. However, most existing GUI agents currently perceive the GUI through text-based representations, such as HTML files or accessibility (a11y) trees \citep{deng2023mind2web, zhou2024webarena, gur2024a}, and interact with the environment by selecting from  predefined list of options or labeled bounding boxes \citep{zheng2024seeact, he2024webvoyager, zhang2024ufo}, similar to the set-of-marks approach described earlier. These representations are often impractical, as they produce excessively long input sequences and can be noisy or incomplete \citep{gou2025navigating}.

Although text-based representations have limitations, multimodal GUI agents require strong grounding performance because errors can lead to failures in executing the task. They also require generalisation capabilities to adapt to different GUIs. As a result, recent efforts have moved away from text-based approaches toward agents that perceive the GUI directly through visual observations and perform pixel-level interactions \citep{cheng2024seeclick, gou2025navigating, shaw2023pixels, hong2024cogagent}, achieving comparable or even superior performance to text-based agents.

Established benchmarks often assume task instructions are specified as step-by-step commands referring explicitly to particular GUI elements and often exhibit limited variability in task design \citep{liu2018reinforcement, bai2021uibert}. More recently, AITW \citep{rawles2023androidinthewild} was introduced, comprising episodes that cover diverse Android versions and devices. Each episode includes a goal described in natural language, the corresponding sequence of actions, and screenshots documenting the interaction. ScreenSpot \citep{cheng2024seeclick} is a GUI grounding evaluation benchmark that 
encompasses mobile, desktop, and web environments. Subsequent versions of \citep{li2025screenspotpro, wu2024atlas} have expanded the domain coverage and incorporated high-resolution inputs tailored to professional settings.
Similarly, OSworld-G \citep{xie2025scaling} provides instructions that address text matching, element recognition, and layout understanding, and it also includes refusal examples, i.e., cases where the instruction does not correspond to any element in the GUI. Finally, two representative works, UGround \citep{gou2025navigating}, and JEDI \citep{xie2025scaling}, provide training resources by pairing referring expressions with GUI screenshots, and supporting both goal-level evaluation and intermediate grounding assessments.

\subsection{Evaluation Metrics}

Evaluation metrics provide standardised criteria for measuring the accuracy, robustness, and generalisation capabilities of vision-language models.

For REC, the commonly employed evaluation metric at the sample-level is the Intersection over Union (IoU) \citep{yu2016modeling, chen2025revisiting} between the predicted and the groundtruth bounding box. 
IoU metric measures the overlap between a predicted bounding box $P$, and its corresponding ground truth bounding box $G$, computed as $IoU = \frac{|P \cap G|}{|P \cup G|}$. This metric captures the geometric properties of the compared objects, including their dimensions and spatial positions, and transforms them into a standardised area-based measurement. 
Since IoU focuses on the proportion of overlap relative to the total area, it provides a scale-independent assessment that remains consistent regardless of the absolute sizes of the objects being evaluated.
Owing to this property, IoU is also applied in RES.
IoU produces a score for an individual sample ranging between 0 and 1, where higher values indicate better alignment between prediction and ground truth. At the dataset-level, performance is often reported as the proportion of predicted results across all test samples that achieve an IoU greater than 0.5.

With regard to generalised expressions, the standard evaluation practice for GREC is to compute Precision@($F_1$=1, IoU$\ge$0.5), defined as the percentage of samples that achieve an $F_1$ score of 1, with the IoU threshold set to 0.5. Given a predicted and ground-truth bounding box, the prediction is classified as a True Positive (TP) if it matches a ground-truth box with an IoU of at least 0.5. Importantly, if multiple predicted boxes match the same ground-truth box, only the prediction with the highest IoU counts as a TP; the remaining matches are considered False Positives (FP). Ground-truth boxes without any matching predictions are counted as False Negatives (FN), and predicted boxes that do not match any ground-truth are classified as FP. The F1 score for each sample is calculated using the standard formula: $F_1 = \frac{2TP}{(2TP + FN + FP)}$. 
A sample is considered successfully predicted if its $F_1$ score equals 1.0. For samples with no ground-truth targets, the $F_1$ score is set to 1 if no bounding boxes are predicted, and to 0 otherwise. The final metric, Precision@($F_1$=1, IoU$\ge$0.5), represents the proportion of samples that meet these criteria. Finally, for GRES, it is common to report the Generalised IoU (gIoU) \citep{Rezatofighietal2019}, defined as the average of all per-image IoU and Complete IoU (cIoU) \cite{zhengetal2019}, which is calculated as the cumulative intersection over the cumulative union of predicted and ground-truth regions.

When evaluating GVQA performance, the evaluation methodology varies depending on whether the question adopts a multiple-choice format with predetermined candidate responses, requires open-ended answer generation, or provides a set of candidate regions. For multiple-choice questions, accuracy is computed straightforwardly as the proportion of correct predictions. For open-ended questions, the evaluation protocol varies across benchmarks. In the pointing subset of Visual-7W, the model is evaluated using standard multiple-choice accuracy. In PointQA\citep{mani2020point},  where answers are open-ended, and the standard protocol involves pre-processing the predicted answer (e.g. removing punctuation and normalising numerical values) before applying string matching against the ground-truth answer. In addition, VizWiz-VQA-Grounding uses both IoU and IoU-per-question to assess grounding performance, while RefChartQA \citep{vogel2025refchartqa} reports localisation Precision@($F_1$=1, IoU$\ge$0.5) as well as a relaxed-accuracy variant, adopted from chart understanding works, which allows a margin of error for floating-point numeric predictions.

Regional descriptions are commonly evaluated using standard image captioning protocols \citep{lin2014microsoft}, with metrics such as BLEU, \citep{papineni2002bleu}, CIDEr \citep{vedantam2015cider}, METEOR \citep{banerjee2005meteor}, and SPICE \citep{anderson2016spice}, which compute n-gram overlaps between the predicted caption and a set of ground-truth captions. On the other hand, datasets designed for holistic captioning are primarily used as a grounding objective during pre-training. 
Consequently, such datasets often use REC or regional descriptions as downstream evaluation tasks. Nonetheless, it is also common to evaluate  Recall@K for grounded holistic captioning by measuring the proportion of mentioned object regions that can be correctly localised in the images.

Captioning conditioned on traces similarly adopts the standard image captioning evaluation protocol. Nevertheless, a well-known limitation of n-gram overlap metrics is that they tend to penalise plausible and semantically correct captions that include novel words not present in the ground-truth references, while favouring captions that use familiar words even if they are less informative. To address these shortcomings, more recent reference-free metrics like CLIPScore and RefCLIPScore \citep{hessel2021clipscore}, compute alignment scores by measuring the cosine similarity between CLIP embeddings \citep{radford2021learning} of the predicted caption and the image, without relying on ground-truth references. This reference-free approach has demonstrated stronger alignment with human judgments and outperforms traditional overlap-based metrics in correlation with human evaluations.

GUI agents are commonly evaluated using goal-oriented and step-wise  metrics. Success rate is the most commonly used protocol for assessing goal-oriented performance. While this type of evaluation is intuitive, it provides only limited insight into agent behaviour and the causes of failure in specific cases. Therefore, it is also common to report the performance of the agent's ability to predict the correct action at each individual timestep. The step-wise performance of an agent depends on the design of the action space, as some actions require only a visual reference, while others also require additional input variables. For example, the ``click'' action only requires specifying a point within the GUI, whereas the ``type'' action also requires providing the textual content to be entered at the visual reference. The visual grounding capabilities of GUI agents are quantified through step-wise measures by computing the percentage of predictions that correctly reference the intended visual region. This is typically done using region-overlap metrics such as IoU, or, in some cases \cite{cheng2024seeclick}, by verifying whether the center of the predicted bounding box falls within the ground-truth region. Lastly, in many cases \citep{rawles2023androidinthewild, li2020mapping}, both goal-oriented and step-wise performance measures are normalised by the task length to allow fair comparison across tasks of different complexities.

% \subsection{Effects of individual components in the recipe of training VLMs}
% \paragraph{Quality of Language Model}
% \paragraph{Quality of Vision Encoder}
% \paragraph{Self vs Cross-Attention}
% \paragraph{Mapping vs Compression}
% \paragraph{Training Regime}

% \section{Approaches for Visual Grounding}\label{sec:approaches}
% % \subsection{Object-Centric Learning}

% \subsection{Grounding in Multimodal Language Modeling}\label{sec:grounding_mllms}

% \subsection{Grounding as a Reasoning Process in Multimodal Language Models}

% \subsection{Reasoning for Grounding in Multimodal Language Models}

%\section{Challenges and Limitations in the Development \& Evaluation of Grounding VLMs}\label{sec:challenges}
\section{Challenges and Future Directions  for Grounding VLM Research}\label{sec:challenges}

Grounding remains a central yet challenging component of VLMs, essential for achieving fine-grained multimodal understanding. Despite significant advances, several obstacles continue to limit the effectiveness, generalisability, and interpretability of these models. In the following, we first examine key challenges inherent to grounding tasks, and then discuss current limitations of existing approaches along with potential directions for future research.

%\subsection{What Matters when Developing Multimodal Grounding Models}
\subsection{Challenges in Grounding VLMs}

The effectiveness of grounding in VLMs is shaped by several intertwined factors, notably the quality of the language and vision encoders, the interplay of self- and cross-attention mechanisms, the balance between mapping and compression, the level of image granularity, and the characteristics of the training regime. The challenges are examined below.

\paragraph{Quality of Language Model \& Vision Encoder} The quality of the unimodal experts is usually determined by their upstream performance.
The base language model is evaluated based on its language modelling \citep{paperno2016lambada, zellers2019hellaswag} capabilities, whereas  instruction-following models are evaluated in terms of their ability to comprehend and execute instructions \citep{hendrycks2021measuring}. On the vision side, the encoder's performance is commonly measured through benchmarks in image classification \citep{deng2009imagenet}, object detection \citep{lin2014microsoft} and image segmentation \citep{caesar2018coco}, which serve as standard indicators of its capabilities.

\textit{Which of the two experts has a greater influence on the performance of the resulting VLM?} Empirical evidence from controlled experiments \citep{laurenccon2024matters}, conducted under fixed numbers of parameters and consistent training configurations, shows that the quality of the language model backbone has a more significant impact on VLM performance than the quality of the vision encoder.

\textit{Are transformer alternatives suitable for Grounding VLMs?} While  transformer-based language models have demonstrated strong performance across many multimodal tasks \citep{zhao2024cobra, qiao2024vl}, their limitations with in-context retrieval \citep{jelassi2024repeat, park2024can} raise concerns about the effectiveness for grounding-specific tasks \citep{pantazopoulos-etal-2023-multitask}.

\paragraph{Self vs Cross-Attention between the Two Modalities} As outlined in \cref{sec:grounding_vlms}, there are two main approaches to combining unimodal representations: (i) the \textit{autoregressive} setting, where the two modalities are concatenated into a single input sequence for the language model; and (ii) the \textit{cross-attention} approach, which interleaves cross-attention blocks within the language model. The autoregressive setup is straightforward and compatible with  modern optimisation techniques \citep{dao2024flashattention}, but imposing a causal structure on patch representations is not entirely intuitive.

In contrast, the cross-attention approach introduces substantially more  parameters through additional backbone components resulting in optimisation difficulties \citep{alayrac2022flamingo, grattafiori2024llama, laurenccon2024matters}. Consequently, the common practice is to freeze the unimodal backbone and only update the parameters of the interleaved cross-attention blocks during training to ensure stability.

\textit{Which of the two approaches to combining representations is more effective?} The authors in \citep{laurenccon2024matters} demonstrate that the cross-attention architecture significantly outperforms the autoregressive approach when the backbone parameters are kept frozen during training. When LoRA \citep{hu2022lora} adapters are introduced in both the vision encoder and the language model, however, the cross-attention architecture underperforms despite its larger parameter count. 

It should be emphasised that these comparisons have been conducted primarily on tasks such as visual question answering, reading comprehension, general knowledge reasoning, and image captioning \citep{goyal2017making, lin2014microsoft, singh2019towards, marino2019ok}. Nonetheless, it is reasonable to suggest that these findings extend to grounding evaluations, as no strong evidence suggests otherwise.

\paragraph{The Effect of Mapping vs Compression and Image Granularity} While compression in image-level understanding may be feasible for many multimodal tasks, performance on perception tasks that require fine-grained understanding such as visual grounding, often benefits from feature-preserving methods. Nevertheless, the choice between mapping and compression is closely tied to the requirements of the downstream application. Consider, for example, GUI agents or agents operating in dynamic environments, which receive a new image observation after each action. Assuming an image resolution of $1120\times1120$ pixels and a standard vision encoder using $16\times16$ tiles will require $(1120/16)^2=4900$ patch embeddings to represent a single image.
A common compromise is to divide the image into several fixed-size sub-images, which are processed independently by the vision encoder, and then employ a mapping connector. In practice, however, sub-images are not independent, and encoding each one separately may lead to a loss of global context. To mitigate this, current strategies include adding both the downscaled full image and the original-resolution image to the list of sub-images, thereby preserving global information.

\paragraph{Training Regime} The development of a VLM is a complex process, akin to following a recipe with many ingredients that must be effectively combined while balancing coarse- and fine-grained granularities across multimodal tasks. Grounding is not merely a by-product of multimodal training but a capability that can be explicitly fostered or inadvertently hindered, depending on how the training regime is designed. Moreover, the multiple stages involved in developing a VLM make such models prone to forgetting previously acquired knowledge or skills due to distribution shifts, task interference, and parameter conflicts \citep{pantazopoulos2024learning, zhai2023investigating, nikandrou2024enhancing, zhou2025learning}.
To address this, it is a common practice to include data from earlier development stages during training to preserve these capabilities \citep{laurenccon2024building}. Taken together, grounding capabilities should be incorporated from the early stages of model development. When incorporating multi-task pre-training objectives, it is also important to consider the learning dynamics associated with task difficulty. For example, a model may only learn to perform visual grounding effectively after first mastering visual question answering. As a result, while continual pre-training may improve the grounding performance, it can also degrade the model's question answering capabilities. This trade-off can be mitigated by carefully selecting data mixtures and sampling distributions \citep{raffel2020exploring}, ensuring that the model converges to a strong starting checkpoint before fine-tuning.

\subsection{Limitations and Opportunities for Further Research}

While grounded VLMs have made significant progress, they still exhibit several limitations that highlight important directions for future research. These limitations and open problems are discussed in detail below.

\paragraph{Lack of Grounding Objectives during Pre-training} Incorporating grounding objectives at all stages of VLM development is crucial. However, only a few studies have explicitly addressed this, for example through grounding captioning \citep{peng2023kosmos}, referring expressions \citep{xue2024xgen} conversations \citep{chen2023shikra, ma2024groma, zhang2024llava}, or GUI elements \citep{gou2025navigating, xie2025scaling}.
Although these approaches provide valuable resources, many rely on pseudo-labels, complex pipelines, or proprietary models \citep{achiam2023gpt}, which can lead to suboptimal quality and raise concerns regarding reproducibility, accessibility, and data contamination. Thus, there is a growing need for publicly available resources with fully documented and transparent dataset creation processes. Future work should draw inspiration from existing VLMs such as Molmo \citep{deitke2024molmo}, which emphasise transparency by thoroughly documenting their data collection procedures.

% Downstream evaluation
\paragraph{Fine-tuning \& Downstream Evaluation} Benchmarks are becoming quickly saturated \citep{kiela2021dynabench, gema2024we}, making it imperative to reconsider their ecological validity \citep{de2020towards}. For instance, established REC benchmarks such as RefCOCO variants have contributed significantly to the evaluation of multimodal models, yet they remain limited in terms of object and vocabulary variability. Although recent efforts have shifted toward more realistic scenarios \citep{he2023grec, chen2025revisiting}, further research in this direction is necessary to ensure that the next generation of grounding VLMs meets usability standards.

Additionally, some benchmarks are derived from the validation or test sets of existing academic datasets. Consequently, models that include such data during supervised fine-tuning will gain an advantage. By contrast, benchmarks should be used to measure model performance, rather than as a hill-climbing training objective.
While fine-tuning on similar examples may improve benchmark scores, it provides limited evidence of a model’s ability to generalize to real-world scenarios. For this reason, it is important for researchers developing VLMs to exclude any images used in the benchmarks they aim to evaluate from their supervised fine-tuning data.

\paragraph{Agents Interacting with Graphical User Interfaces} This field remains in its nascent stages, with ongoing efforts focused on developing models, datasets and evaluation metrics.
% Many works on high- and low-level action control are policies derived from VLMs and therefore any potential advances in their grounding capabilities directly benefits such agents \citep{zhou2025chatvla, shukor2025smolvla}. 
Recent work has shown that GUI agents with multimodal grounding capabilities can match or even surpass the performance of similar agents that employ text-based representations \citep{cheng2024seeclick, gou2025navigating,  hong2024cogagent, shaw2023pixels}. This demonstrates the potential of such agents for applications that interact with web, operational systems or mobile devices through VLMs.

\paragraph{Verifying Architectural Design Choices for Grounding VLMs} In \cref{sec:grounding_vlms}, we identified key elements in the design of modern VLMs that influence their performance \citep{laurenccon2024matters, karamcheti2024prismatic, laurenccon2024building, lin2024preserve}.
Although these studies do not explicitly examine grounding capabilities, we expect that their findings will be applicable to grounding tasks.
Nevertheless, further investigation into the design space of VLMs is necessary to validate design choices in the context of grounding.

\paragraph{Multimodal Grounding \& Reasoning} 
There are several promising future directions inspired by the final stage of grounding VLM development, in which the model learns directly from feedback on its outputs. Visual grounding relies on the model's ability to understand spatial relationships, orientation, and the underlying semantics of entities in the image. Accordingly, multimodal chain-of-thought and reasoning paradigms have the potential to enhance these models' reasoning capabilities by generating explanations and ensuring the factual accuracy of their reasoning traces. Intermediate reasoning steps that bridge visual perception and semantic understanding not only provide transparency regarding the model's grounding decision but may also improve performance in challenging scenarios involving occlusion, ambiguous spatial relationships, or multi-step visual reasoning. Such approaches could enable VLMs to address applications in which the reasoning process is as important as the final output. The factuality of these reasoning traces is crucial not only for model interpretability but also for fostering trust in high-stakes applications, where understanding the model's decision-making process is essential for validation and error correction.

\section{Conclusion}
This survey has explored the evolving landscape of visual grounding in modern VLMs, underscoring its central role in bridging vision and language to enable fine-grained multimodal understanding.  We presented a comprehensive review of the development, applications, and evaluation of grounded VLMs across key domains, including referring expression comprehension, grounded visual question answering, grounded captioning, and GUI agents. Our analysis highlighted the architectural design choices and training regimes that underpin VLMs, with a particular attention to how grounding capabilities are shaped by factors such as pixel-level representations, multimodal connectors, and the quality of language backbone. Finally, we discussed promising directions for further research on grounded VLMs, encompassing the development of more effective pre-training data, the design of downstream applications, the architectural design aspects for the next-generation VLMs, and the integration of grounding with multimodal reasoning.

% \section{Examples of citations, figures, tables, references}
% \label{sec:others}
% Citations... \cite{kour2014real,kour2014fast}. 
% \begin{quote}
%   Hasselmo, et al.\ (1995) investigated
% \end{quote}

% \subsection{Figures}

% See \cref{fig:fig1}. Here is how you add footnotes. \footnote{Sample of the first footnote.}

% \begin{figure}
%   \centering
%   \fbox{\rule[-.5cm]{4cm}{4cm} \rule[-.5cm]{4cm}{0cm}}
%   \caption{Sample figure caption.}
%   \label{fig:fig1}
% \end{figure}

% \begin{figure} % picture
%     \centering
%     \includegraphics{test.png}
% \end{figure}

% \subsection{Tables}

% See Table~\ref{tab:table}.

% \begin{table}
%  \caption{Sample table title}
%   \centering
%   \begin{tabular}{lll}
%     \toprule
%     \multicolumn{2}{c}{Part}                   \\
%     \cmidrule(r){1-2}
%     Name     & Description     & Size ($\mu$m) \\
%     \midrule
%     Dendrite & Input terminal  & $\sim$100     \\
%     Soma     & Cell body       & up to $10^6$  \\
%     \bottomrule
%   \end{tabular}
%   \label{tab:table}
% \end{table}

% \subsection{Lists}
% \begin{itemize}
% \item Lorem ipsum dolor sit amet
% \item consectetur adipiscing elit. 
% \item Aliquam dignissim blandit est, in dictum tortor gravida eget. In ac rutrum magna.
% \end{itemize}

\bibliographystyle{unsrt}  
\bibliography{references}  %%% Remove comment to use the external .bib file (using bibtex).
%%% and comment out the ``thebibliography'' section.

%%% Comment out this section when you \bibliography{references} is enabled.
% \begin{thebibliography}{1}

% \bibitem{kour2014real}
% George Kour and Raid Saabne.
% \newblock Real-time segmentation of on-line handwritten arabic script.
% \newblock In {\em Frontiers in Handwriting Recognition (ICFHR), 2014 14th
%   International Conference on}, pages 417--422. IEEE, 2014.

% \bibitem{kour2014fast}
% George Kour and Raid Saabne.
% \newblock Fast classification of handwritten on-line arabic characters.
% \newblock In {\em Soft Computing and Pattern Recognition (SoCPaR), 2014 6th
%   International Conference of}, pages 312--318. IEEE, 2014.

% \bibitem{hadash2018estimate}
% Guy Hadash, Einat Kermany, Boaz Carmeli, Ofer Lavi, George Kour, and Alon
%   Jacovi.
% \newblock Estimate and replace: A novel approach to integrating deep neural
%   networks with existing applications.
% \newblock {\em arXiv preprint arXiv:1804.09028}, 2018.

% \end{thebibliography}

\end{document}